\pdfoutput=1

\documentclass[11pt]{article}

\usepackage[final]{acl}
\usepackage{times}
\usepackage{latexsym}
\usepackage[most]{tcolorbox}
\usepackage{subcaption}
\usepackage{graphicx}
\tcbset{colback=gray!3, colframe=gray!50, boxrule=0.5pt, arc=2pt}
\usepackage{xcolor}
\usepackage{array}
\usepackage{tabularx}
\usepackage{multirow}
\usepackage{listings}
\usepackage{tabularx}
\usepackage{booktabs}

\definecolor{domaincol}{HTML}{1F78B4} 
\definecolor{trigcol}{HTML}{E31A1C}   
\definecolor{etypecol}{HTML}{FF7F00}  
\definecolor{argcol}{HTML}{33A02C}    

\usepackage{booktabs}
\usepackage{graphicx}
\usepackage{amssymb}
\usepackage{xcolor}
\usepackage[table]{xcolor}

\usepackage[T1]{fontenc}

\usepackage[utf8]{inputenc}

\usepackage{microtype}

\usepackage{inconsolata}

\usepackage{graphicx}

\title{A Multi-Domain and Multi-Task Generative Framework with Explicit Task and Domain Conditioning for Cross-Domain Event Extraction}

\author{
Siting Liang$^{1,2}$
\and
Omar Adjali$^1$\and
Daniel Sonntag$^{1,2}$\\
$^1$German Research Center for Artificial Intelligence\\
$^2$Carl von Ossietzky University of Oldenburg\\
\{siting.liang, omar.adjali, omair\_shahzad.bhatti, 
daniel.sonntag\}@dfki.de}

\begin{document}
\maketitle
\begin{abstract}
Event extraction aims to identify event triggers, classify event types, and extract arguments to construct structured event representations. Despite strong in-domain performance, developing models that generalize robustly across domains remains challenging due to variations in contextual expressions and event schemas. Prior unified and multi-task approaches improve in-domain accuracy but exhibit limited flexibility when applied to unseen domains. Even large language model–based methods that provide full event ontologies at inference time often underperform compared to smaller, task-specific fine-tuned models. We propose a unified multi-domain and multi-task training framework that models heterogeneous event schemas within a single model. Our approach introduces domain conditioning signals, jointly with task-specific prompts, enabling dynamic adaptation to dataset-specific schemas without requiring complete event label sets at inference time. The framework supports both pipeline and end-to-end extraction settings, facilitating efficient task- and domain-level transfer. Experiments on diverse event extraction benchmarks demonstrate that our method achieves competitive performance, strong cross-domain generalization, and practical scalability, while preserving domain-specific precision.
\end{abstract}

\section{Introduction}
\textbf{Event extraction (EE)} is a fundamental task in information extraction that aims to identify and structure mentions of events from unstructured text. 
 A typical event extraction pipeline involves two core tasks \cite{yang-etal-2019-exploring}: (i) \textbf{Event Detection (ED)},  which detects spans of text that evoke the occurrence of an event and assigns the detected trigger word to a predefined event type; and (ii) \textbf{Event Argument Extraction (EAE)}, which identifies and labels the participants or entities that fulfill specific roles in the event. Together, these tasks enable the construction of structured representations of events that support downstream applications such as event-centric knowledge graph construction \cite{gottschalk2019eventkg}. 

\begin{figure}[t]
    \centering
    \includegraphics[width=\linewidth]{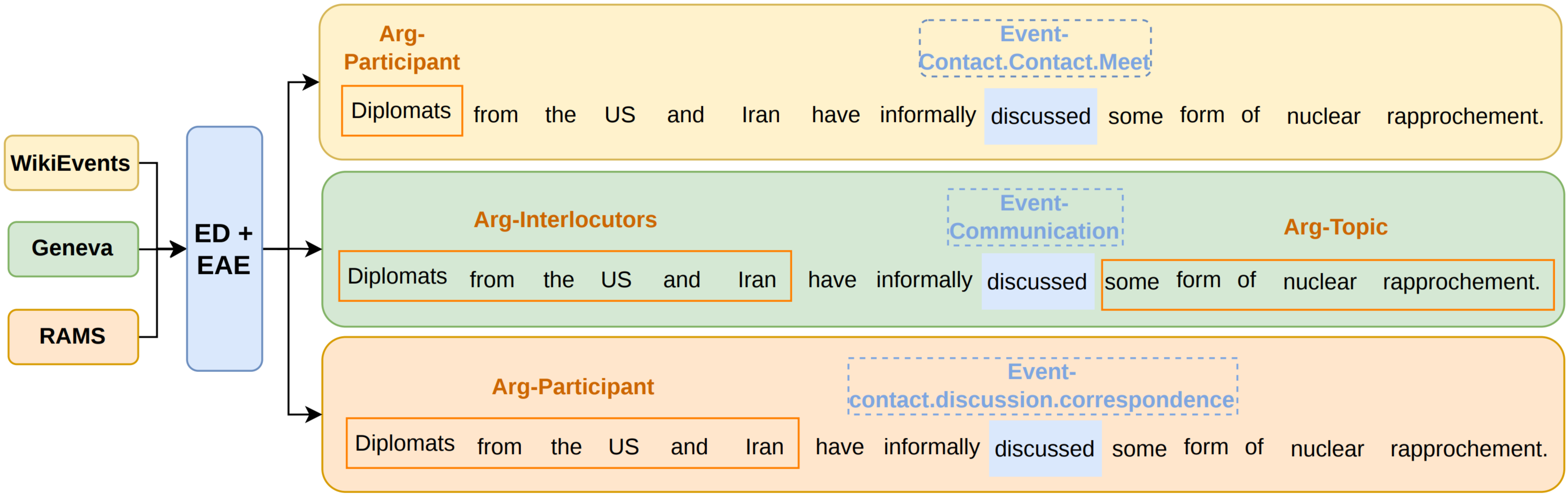}
    \caption{Illustration of cross-domain event extraction, where a single unified model generates domain-specific predictions for the same input text by conditioning on the target domain at inference time.}
    \label{fig:ee_example}
\end{figure}

\begin{figure*}
   \centering
    \includegraphics[width=\linewidth]{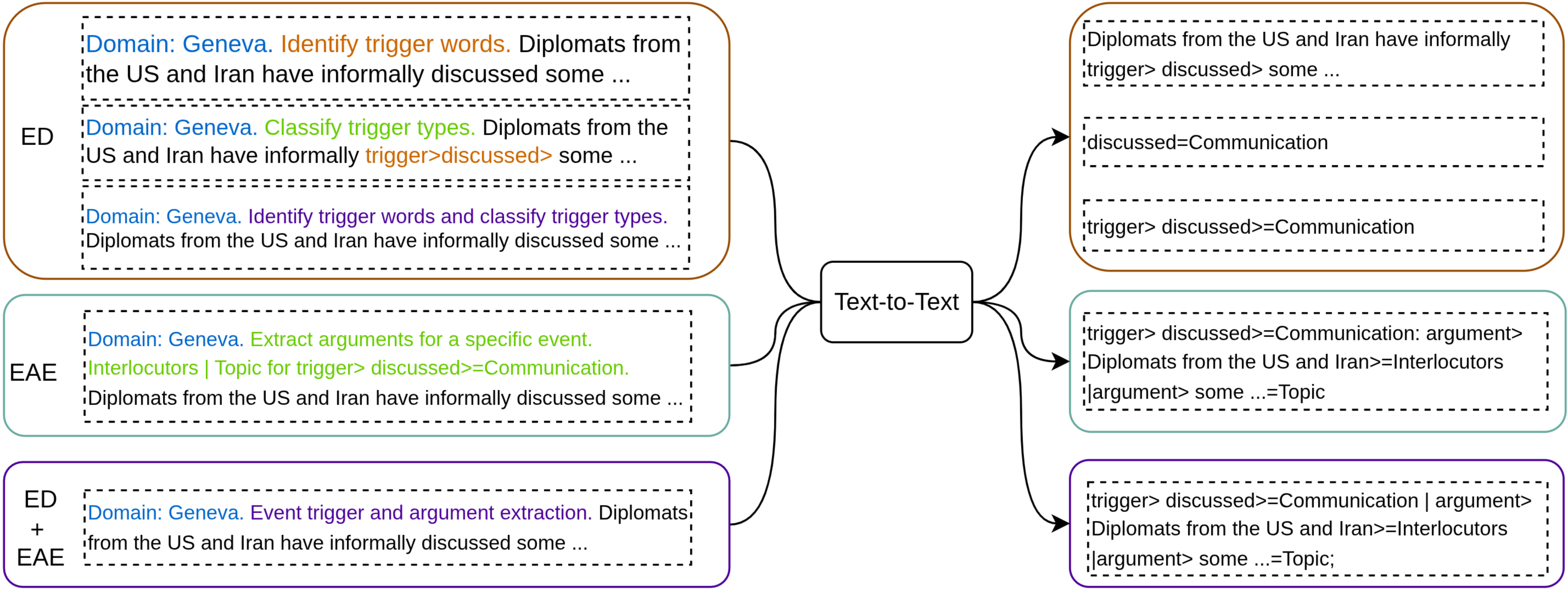}
    \caption{The unified framework is built on a text-to-text architecture, allowing a single model to learn multiple tasks simultaneously through task-specific prompts while training on multiple datasets. During training, domain-conditioned signals (e.g., \textcolor{blue}{Domain: Geneva.}) enable robust cross-domain transfer while maintaining strong domain-specific precision.}
    \label{fig:ee_pipeline} 
\end{figure*}

 Prior work on event extraction predominantly relies on discriminative models that decompose the task into separate modules, or on unified generative models that jointly encode event structures within a single framework \cite{Huang23textee,wang2023instructuie}. Building on the latter paradigm, we extend unified event extraction to a fully multi-domain and multi-task setting, training a single model to handle all domains and event subtasks in a scalable and efficient manner. Figure~\ref{fig:ee_example} illustrates cross-domain event extraction at inference time, where our unified multi-domain fine-tuned model leverages domain indicators (e.g., \textit{WikiEvents}, \textit{Geneva}, \textit{RAMS}) to generate domain-specific predictions given the identical input text. 

We adopt a text-to-text multi-task framework that reformulates core event extraction subtasks into a unified generation paradigm, enabling a single encoder–decoder model to jointly perform event detection (ED) and event argument extraction (EAE) \cite{lu-etal-2022-unified}. This formulation builds on prior work in prefix-based conditioning, controllable generation, and unified information extraction, where task-specific tokens serve as lightweight prompts to guide generation across different subtasks \cite{Hsu23tagprime,Hsu23ampere}. Extending this paradigm, we further generalize the framework to a multi-domain setting by incorporating additional domain-level conditioning. Specifically, lightweight domain indicators are prepended to the input sequence (e.g., \textit{Domain: Geneva.}), allowing the model to adapt its generation behavior to diverse domain-specific event patterns without requiring explicit schema definitions or full event ontologies at inference time. As illustrated in Figure~\ref{fig:ee_pipeline}, the resulting framework unifies multi-task and multi-domain learning under a single generative model guided by both task-specific prompts and domain indicators. The unified framework allows a single model to operate flexibly in either a pipeline setting, detecting and classifying triggers and arguments extraction in successive steps: \textbf{Pipeline} of \textbf{ED} $\Rightarrow$ \textbf{EAE}, or in an \textbf{End-to-End} manner that jointly generates structured event representations (\textbf{ED+EAE}). Detailed descriptions of the unified framework and both operating modes are provided in the Section~\ref{sec:approach}. 

Our experiments demonstrate that a single T5-base model \cite{flan_t5_base} can effectively handle all event extraction subtasks, operating either sequentially in a pipeline setting or jointly through end‑to‑end prediction.  The key contributions of our work are: (1) proposing a single generative model capable of flexible event extraction in multiple paradigms, (2) demonstrating mutual benefits of joint multi-task and multi-domain training, and (3) showing effective cross-domain generalization with a single model. We will publish our code and T5-based event extraction models of this work. 

\section{Related Work}
Event extraction systems have traditionally relied on discriminative sequence-labeling approaches, where classifiers are fine-tuned on human-annotated label sets, often resulting in domain-specific models trained separately for each dataset \cite{Wadden19dygiepp, yang-etal-2019-exploring}. Recent mainstream event extraction systems are typically built on joint or unified modeling paradigms, where entities, relations, and events are learned together through multi-task training to boost in-domain performance, especially under limited training resources \cite{Lin20oneie,lu-etal-2022-unified, Hsu22degree, Hsu23tagprime}. However, these approaches remain tightly coupled to fixed label sets and predefined schemas, which limits their scalability and transferability across domains. Moreover, they are often evaluated on only a small number of event extraction datasets, leaving their generalization to diverse domains and heterogeneous event schemas largely unvalidated \cite{huang2024textee}. 

Zero-shot prompting with large language models (LLMs) suggests the promise of generalized event extraction, but in practice yields consistently low performance and relies heavily on explicitly provided schemas, labels, and output constraints, and still exhibit weak schema adherence. Instruction fine-tuning can improve schema adherence and reduce format inconsistency and invalid outputs \cite{wang2023instructuie,zuo2025rule}. However, these approaches scale poorly to evolving schemas and provide limited domain transferability. While constraint-aware instructions and optimization improve output validity, they focus on generation correctness rather than representation transfer, leaving LLM-based methods behind fine-tuned non-LLM models in F1 performance \cite{srivastava2025instruction, adjali2026aligning}. 

\section{Approach}
\label{sec:approach}
 We propose a unified event extraction framework that balances performance, efficiency, and flexibility across tasks while scaling seamlessly to a number of diverse domains. As illustrated in Figure~\ref{fig:ee_pipeline}, our approach formulates event extraction as a prompt-conditioned text-to-text generation problem, where each input $S$ is prefixed with a domain indicator $\texttt{[Domain: } d \texttt{]}$ and a task-specific prompt $\pi_t$.  The domain indicator $d$ is a simple string (e.g., \textit{Geneva}, \textit{RAMS}) that provides lightweight contextual signals about domain-specific event semantics. The task prompt $\pi_t$ is defined using fixed natural language templates, each corresponding to a distinct subtask (e.g., trigger identification, trigger classification, argument extraction), enabling a unified model to handle multiple tasks within a shared framework.

\paragraph{Multi-domain Multi-task Learning.} Our framework extends prior unified formulations by jointly addressing multiple tasks and domains, enabling a single model to learn shared structures while adapting to heterogeneous event semantics. Training minimizes the negative log-likelihood (\ref{eq:learning})jointly across tasks and domains, enabling a unified model to perform multi-task, multi-domain event extraction. 
\begin{equation}
\resizebox{\columnwidth}{!}{$
p_{\theta}(y \mid d, \pi_t, S)
= \prod_{i=1}^{|y|}
p_{\theta}(y_i \mid y_{<i}, d, \pi_t, S)
$}
\label{eq:learning}
\end{equation}

\paragraph{Multi-task Formulation.}
Table~\ref{tab:abstract_patterns} summarizes the input–target patterns used in our unified multi-task learning framework. We design structured target sequences to support different modeling configurations, including pipeline and end-to-end settings. This diverse multi-task formulation, on the one hand, supports flexible operation modes by unifying pipeline and end-to-end configurations within a single framework. On the other hand, it augments training data through heterogeneous input–target patterns, enabling the model to capture a broader range of structural dependencies, from span-level trigger prediction and type classification to schema-guided argument extraction and fully structured event generation, thereby improving generalization and robustness across domains and tasks.

\begin{table}[t]
\centering
\resizebox{\columnwidth}{!}{
\begin{tabular}{l l l}
\toprule
\textbf{Task(Mode)} & \textbf{Task-Specific Input Pattern} & \textbf{Target Pattern} \\
\midrule
ED (P1) 
& \texttt{[Domain:$d$][Task: TI] $S$}
& \texttt{ $S$ + <trigger> $w$ >}
\\
ED (P2) 
& \texttt{[Domain:$d$][Task: TC] $S$ + <trigger>}
& \texttt{$w$ = $t$}
\\
ED (E2E) 
& \texttt{[Domain:$d$][Task: TI+TC] $S$}
& \texttt{<trigger> $w$ >=$t$}
\\
\midrule
EAE (P) 
& \texttt{[Domain:$d$][Task: EAE] $S$ + ($w$, $t$)}
& \texttt{$a_1=r_1$ ; $a_2=r_2$}
\\
\midrule
ED+EAE (E2E) 
& \texttt{[Domain: $d$][Task: EE] $S$}
& \texttt{$w=t$ | $a_1=r_1$ ; $a_2=r_2$}
\\
\bottomrule
\end{tabular}
}
\caption{Abstract input–output patterns for multi-task event extraction under pipeline (P) and end-to-end (E2E) settings. Symbols denote: $w$ (trigger word), $t$ (event type), $a$ (argument text), and $r$ (role type).}
\label{tab:abstract_patterns}
\end{table}

\paragraph{Pipeline Event Generation. }

In pipeline mode, event detection is decomposed into sequential subtasks: trigger identification (ED–P1), type classification (ED–P2), and argument extraction (EAE–P). Concretely, the model first predicts triggers using
\textit{[Domain:$d$][Task: TI] $S$}, then classifies their types via
\textit{[Domain:$d$][Task: TC] $S$ +<trigger> }, and finally performs argument extraction with
\textit{[Domain:$d$][Task: EAE] $S$ + ($w$, $t$)}.
A key advantage of this formulation is that each input corresponds to \textit{a single event instance}, conditioning argument extraction on \textit{a specific trigger and event type}. This allows EAE to leverage schema-constrained prompts instantiated from the predicted type and domain, explicitly encoding valid argument roles and guiding the model to extract event-specific arguments with higher precision. However, this design is sensitive to error propagation: incorrect trigger or type predictions can lead to entirely incorrect downstream arguments.
To mitigate this, trigger identification and type classification can also be performed jointly in an end-to-end formulation (ED–E2E) using
\textit{[Domain:$d$][Task: TI+TC] $S$}, which serves as an additional subtask during multitask training, increases training diversity, and provides greater flexibility at inference time. Argument extraction is executed based on the results of either \textbf{ED–pipeline} or \textbf{ED–E2E}.

\paragraph{End-to-End Event Generation. }
The model also supports a fully end-to-end event extraction mode (ED+EAE–E2E) with
\texttt{[Domain:$d$][Task: EE] $S$}, directly generating complete event structures in a single pass without intermediate predictions, thereby reducing cascading errors. However, this approach does not explicitly incorporate schema constraints for EAE, requiring the model to implicitly capture role structures and dependencies, which poses additional challenges for accurate and consistent generation.

\section{Experiments}
\label{sec:experiments}
\paragraph{Datasets.}
We conduct our main experiments on publicly accessible event extraction datasets. For standardized evaluation, we adopt the preprocessed versions provided by \citet{huang2024textee}, excluding other information extraction datasets that lack event argument annotations and firstly focusing on English-only datasets. The datasets employed in this study are as follows:

\begin{itemize}
\item \textbf{CASIE} \citep{CASIE_Satyapanich_Ferraro_Finin_2020}: A cybersecurity event extraction dataset. It defines event subtypes (e.g., Vulnerability Disclosure, Attack) focusing on the critical domain of cybersecurity intelligence.
\item \textbf{Geneva} \citep{geneva_parekh2023geneva}: A large-scale benchmark for event extraction aligned to FrameNet \cite{fillmore-etal-2002-framenet} structures. It specifies hundreds of event types and roles across diverse textual sources, including books, articles, journals, and Wikipedia.  
\item \textbf{Genia2013} \citep{kim-etal-2013-genia}: A widely used biomedical event extraction dataset, it centers on molecular-level events (e.g., Gene Expression, Phosphorylation) in scientific literature, representing a challenging domain characterized by specialized terminology.
\item \textbf{M2E2} \citep{m2e2_li2020cross}: A multimedia news-domain event extraction dataset. It defines events and arguments from annotated news articles and images, bridging textual and visual modalities. Only the text-based annotations for events and arguments are applied in our experiments. 
\item \textbf{RAMS} \citep{rams_ebner2020mult}: A news-domain event extraction dataset. It defines argument roles across multiple sentences, emphasizing cross-sentence argument linking in news articles.
\item \textbf{WikiEvents} \citep{wikievents}: A Wikipedia current-events domain event extraction dataset. It defines event triggers and arguments at the document level. It requires identifying event triggers and their arguments throughout an entire document.

\end{itemize}
Examples of event and argument roles are shown in Table~\ref{tab:dataset}.
\begin{table}[h!]
\centering
\renewcommand{\arraystretch}{1.3}
\resizebox{\columnwidth}{!}{
\begin{tabular}{|p{2.5cm}|p{4cm}|p{8.5cm}|}
\hline
\textbf{Dataset} & \textbf{Example Event Type} & \textbf{Example Argument Roles} \\
\hline

CASIE \newline (Cybersecurity)  & Attack\_Databreach & 
\textit{Attacker}, \textit{Target}, \textit{Malware}, \textit{Vulnerability}, \textit{Compromised Data}, \textit{Instrument} \\  
\hline

Geneva \newline  (General) & Protest & 
\textit{Protester}, \textit{Target}, \textit{Place}, \textit{Time}, \textit{Reason}, \textit{Organizer} \\ 
\hline

Genia2013 \newline (Biomedical) & Gene\_expression & 
\textit{Theme (Gene/Protein)}, \textit{ Agent (Cause)}, \textit{Site (Location)}, \textit{Resulting Entity} \\ 
\hline

M2E2 \newline (Geopolitical) & Conflict\_attack  & 
 \textit{Location}, \textit{Victims}, \textit{Casualties}, \textit{Time} \\ 
\hline

RAMS \newline (Newswire) & Contact\_negotiate\_meet & 
\textit{Participant}, \textit{Location}, \textit{Time}, \textit{Topic}, \textit{Outcome} \\ 
\hline

WikiEvents \newline (Wikipedia) & Personnel\_EndPosition & 
\textit{Entity}, \textit{Organization}, \textit{Position}, \textit{Place}, \textit{Time}, \textit{Reason} \\ 
\hline

\end{tabular}
}
\caption{Representative event-type examples and their domain-specific argument roles across diverse event-extraction datasets.}
\label{tab:dataset}
\end{table}

Table~\ref{tab:data-statistics} summarizes the dataset splits, number of samples, annotated event types, argument roles, and annotation instances. These datasets exhibit substantial variation in both scale and semantic complexity. For instance, \textit{Geneva} and \textit{RAMS} contain large and diverse event ontologies with over 100 event types and tens to hundreds of role types, while \textit{CASIE} and \textit{M2E2} involve fewer event types but differ significantly in data density and annotation sparsity. 
\begin{table}[h!]
\centering
\resizebox{\columnwidth}{!}{
\begin{tabular}{lcccccc}
\toprule
\textbf{Dataset} & \textbf{Split} & \textbf{Event Types} & \textbf{Event Mentions} & \textbf{Role Types} & \textbf{Arguments} \\
\midrule
CASIE      & Train(1047) & 5   & 5980  & 26  & 15869  \\
          & Dev(218)  & 5   & 1221  & 26  & 3175   \\
           & Test (218)  & 5   & 1268  & 26  & 3531   \\
\midrule
Geneva     & Train (2582) & 115 & 5290  & 220 & 8618  \\
          & Dev (509)  & 115 & 1016  & 159 & 1683    \\
           & Test (593)  & 115 & 1199  & 171 & 2013   \\
\midrule
Genia2013  & Train (420)  & 13  & 4077  & 7   & 3921   \\
          & Dev (105)  & 10  & 950   & 7   & 858    \\
           & Test(139)  & 11  & 974   & 7   & 881    \\
\midrule
M2E2       & Train (4211) & 8  & 748   & 15  & 1120   \\
          & Dev (901)  & 8  & 183   & 15  & 280    \\
           & Test (901)  & 8  & 174   & 15  & 259    \\
\midrule
RAMS       & Train (7827) & 139 & 7287 & 65  & 16951  \\
          & Dev (910)  & 136 & 910  & 64  & 2132   \\
           & Test (910)  & 135 & 910  & 63  & 2123   \\
\midrule
WikiEvents & Train (450)  & 50  & 3131 & 57  & 4393   \\
           & Dev (53)   & 39  & 422  & 43  & 592    \\
           & Test (62)   & 38  & 379  & 46  & 516   \\
\bottomrule
\end{tabular}
}
\caption{Statistics of six event extraction datasets applied in our experiments.}
\label{tab:data-statistics}
\end{table}

In addition, the number of event mentions and arguments also varies widely across datasets: \textit{RAMS} and \textit{CASIE} include dense annotations with thousands of event instances and arguments, whereas \textit{M2E2} contains relatively sparse event mentions despite a larger number of training samples. Additionally, \textit{Genenia2013} focuses on specialized biomedical events with a limited role schema (7 role types), in contrast to the more complex role structures observed in \textit{Geneva} (more than 200 role types).  
Overall, this collection reflects a wide spectrum of domain shifts in event semantics, label distributions, and structural complexity, making it well-suited for evaluating cross-domain event extraction.

\begin{table}[ht]
\centering
\resizebox{\columnwidth}{!}{
\begin{tabular}{lrrr}
\toprule
Dataset & \#Samples & Mean \#Tokens & Mean \#Annotated Tokens  \\
\midrule
CASIE      & 1047 & 295.2 & 46.77   \\
Geneva     & 2582 & 26.6  & 14.08  \\
Genia2013  & 420  & 225.2 & 20.92 \\
M2E2       & 4211 & 28.2  & 0.520   \\
RAMS       & 7287 & 132.9 & 5.416   \\
WikiEvents & 450  & 329.0 & 18.92 \\
\bottomrule
\end{tabular}
}
\caption{Overview of token annotation statistics. The first column shows the number of annotated samples, the second column indicates the average number of tokens per sample, and the third column reports the average number of tokens with assigned labels.}

\label{tab:dataset-overview}
\end{table}
The datasets exhibit substantial variation in sequence length and annotation density across domains. WikiEvents and CASIE contain long articles as input, with average lengths of 329.0 and 295.2 tokens, respectively, followed by Genia2013 (225.2) and RAMS (132.9). In contrast, Geneva and M2E2 consist of sentence input, averaging only 26.6 and 28.2 tokens per sample.
However, the number of annotated tokens does not scale proportionally with input length, revealing notable sparsity in several datasets. Geneva has the highest annotation density, nearly half of the input tokens are labeled per sample. Genia2013 also exhibit moderate annotation coverage (14.08 and 20.92, respectively). In contrast, datasets such as RAMS (5.416), WikiEvents (18.92), and especially M2E2 (0.520) show significantly fewer annotated tokens relative to their input length. This indicates that, for these datasets, only a small fraction of the text is labeled, resulting in highly sparse supervision.
Overall, while some datasets provide relatively dense annotations,  M2E2 and RAMS are characterized by sparse labeling. These pose additional challenges for cross-domain learning and evaluation. 

\paragraph{Models.} We conducted preliminary experiments (see Appedix~\ref{sec:preliminary}) to identify a suitable base model in our experiments. Similar to \citet{chanin2023opensource}, we found that \textbf{T5-Base} \cite{flan_t5_base} demonstrated the best balance between performance and computational efficiency. Scaling up to \textbf{T5-Large} provided negligible improvement in performance relative to the increased computational cost. While \textbf{BART-base} \cite{lewis2020bart} and \textbf{GPT-2} \cite{radford2019language_gpt2} achieved competitive results in trigger identification task, they exhibited instability in argument extraction, consistent with prior observations in related studies \cite{Li21bartgen, chanin2023opensource}. Hence,  \textbf{T5-Base} is selected as the primary architecture in the multi-domain experiments. 

\paragraph{Training.} We train six single-domain models and one multi-domain joint model. Each model is trained for 40 epochs, enabling a controlled comparison between in-domain specialization and cross-domain generalization. All models are based on \texttt{T5ForConditionalGeneration}, optimized with a learning rate of \texttt{5e-05}. We apply gradient accumulation with \texttt{gradient\_accumulation\_steps=4} and clip gradients with a maximum norm (\texttt{max\_grad\_norm=1.0}) to ensure stable training.

\paragraph{Post-processing.}
Given the generative nature of our framework, the model outputs linearized text sequences encoding triggers, event types, and arguments. To recover structured event representations, we apply a deterministic post-processing procedure that parses these sequences into structured records. Specifically, we first extract trigger spans and their associated event types from patterns such as \textit{ $w$ > = $t$}. Based on the identified triggers, we then parse argument tuples by matching patterns of the form \textit{$a$=$r$}, optionally conditioned on trigger-specific context in pipeline settings. Each event is represented by a trigger (with text span, type, and offset) and a set of associated arguments (with text, role, and offset). The final outputs are organized in a hierarchical format grouped by inference mode (pipeline or end-to-end), as illustrated below:
\begin{lstlisting}[basicstyle=\ttfamily\footnotesize, breaklines=true]
{
  "events": {
      "pipeline": [
        {
          "trigger": {
            "text": "{w}",
            "type": "{t}",
            "offset": [int, int]
          },
          "arguments": [
            {
              "text": "{a}",
              "role": "{r}",
              "offset": [int, int] 
            }
          ]
        }
      ],
      "ened-to-end": [...]
    }
  }
}
\end{lstlisting}
This post-processing step is required to bridge the gap between free-form text generation and structured event extraction, ensuring consistent and comparable evaluation across different tasks and inference modes.

\paragraph{Evaluation.}
For event detection (ED), trigger identification (TI) evaluates the correctness of predicted trigger offsets, while trigger classification (TC) additionally requires correct event types. For event argument extraction (EAE), following \citet{Huang23textee}, argument identification (AI) evaluates correct argument spans linked to triggers, and argument classification (AC) further requires correct argument roles. Both AI and AC enforce correct argument--trigger associations, enabling faithful evaluation of complete event structures.

We evaluate TI, TC, AI, and AC using precision, recall, and F1 based on matching between predicted(p) and gold structures(g):

\begin{itemize}
    \item \textbf{TI}: match trigger spans $(w_p, w_g)$.
    \item \textbf{TC}: match trigger span and event type $((w_p, t_p), (w_g, t_g))$.
    \item \textbf{AI}: match argument--trigger pairs $((a_p, w_p), (a_g, w_g))$.
    \item \textbf{AC}: match argument, role, and trigger $((a_p, r_p, w_p), (a_g, r_g, w_g))$.
\end{itemize}

\section{Results}
We report the general results for TI, TC, AI and AC in Table~\ref{tab:f1_group1} and Table~\ref{tab:f1_group2} after 40 epochs of training. 
Models are evaluated in different modes, including ED-gold (gold triggers), ED-pipeline, ED-e2e followed by EAE in pipeline mode,  and fully end-to-end generation. A clear performance gap between \textbf{ED-gold} and all predicted-trigger settings confirms that trigger identification errors remain the primary bottleneck for event extraction. This is particularly evident in the consistent drop from TC to AI/AC under both pipeline and end-to-end settings, underscoring the critical role of accurate trigger localization.
We also include comparisons with representative prior single-domain systems.

\paragraph {Benefit of Multi-domain Learning.} 
jointly training on all datasets achieves performance comparable to (and often exceeding) single-domain models, while using a \emph{single unified model} across all datasets. This provides a clear practical advantage, as it eliminates the need to maintain and deploy separate models for each domain, significantly improving scalability and efficiency. Empirically, the benefits are most evident on resource-limited or schema-diverse datasets such as Genia2013 and WikiEvents. For example, multi-domain training leads to substantial gains in TC, AI, and especially AC on Genia2013, where argument classification improves markedly compared to single-domain training. Similarly, on WikiEvents, both trigger and argument metrics consistently increase under the multi-domain setting. These results indicate that knowledge learned from high-resource domains (e.g., CASIE, M2E2, Geneva) transfers effectively to lower-resource or structurally different domains. Moreover, the gains are more pronounced for argument-level metrics (AI, AC) than for trigger-level metrics (TI, TC). This suggests that argument extraction benefits more from shared cross-domain representations, as argument roles and semantic patterns generalize across datasets, even when trigger vocabularies differ. In contrast, trigger identification remains more domain-specific and thus shows smaller improvements.

As shown in Figures~\ref{fig:epochs_casie}--\ref{fig:epochs_wikievents} from Appendix~\ref{app:multi-domain learning}, joint multi-domain training consistently yields stable improvement across epochs. Precision stabilizes early and remains largely unchanged, while recall improves progressively with training, with the most substantial gains observed in most domains. These observations indicate that shared representation primarily enhances coverage rather than classification confidence. This trend is especially pronounced for end-to-end (E2E) mode. 

\paragraph{ED-pipeline, ED-e2e vs. fully End-to-End.} 
Both ED-pipeline and ED-e2e consistently outperform the fully generative end-to-end setting, particularly on AI and AC. While the gap on TI and TC is relatively small, the difference becomes substantial for argument extraction. Moreover, ED-e2e achieves comparable or slightly better trigger performance than ED-pipeline, while maintaining strong argument scores, demonstrating the advantage of jointly modeling TI and TC under structured supervision. While ED-e2e jointly predicts TI and TC, it still leverages \emph{schema-constrained decoding} for downstream arguments prediction, where valid argument roles are conditioned on the predicted event type. In contrast, the fully end-to-end setting must generate triggers, arguments, and roles simultaneously without explicit constraints. The fully end-to-end mode often fails to predict valid or consistent argument roles, leading to significant degradation in AC. This trend highlights the advantage of the pipeline mode to incorporate structured schema knowledge during argument prediction.

\begin{table*}[h!]
\centering
\resizebox{\textwidth}{!}{
\begin{tabular}{l|cccc|cccc|cccc}
\toprule
 & \multicolumn{4}{c|}{CASIE} & \multicolumn{4}{c|}{M2E2} & \multicolumn{4}{c}{Genia2013}  \\

Metrics & TI & TC & AI & AC & TI & TC & AI & AC & TI & TC & AI & AC  \\
\midrule
 \multicolumn{13}{c}{\textit{Single-Domain Multi-Task Training}}  \\
\hline
ED-gold & -- & 95.1 & 65.1 & 61.6 & -- & 88.8 &58.4 & 51.5& -- & 67.9 & 66.7 & 51.9 \\
ED-pipeline & 67.7 & 66.2 & 56.4 & 52.9 & 68.2 & 62.4 &  53.8 & 40.4& 41.6 & 33.4 & 43.8 & 31.5 \\
ED-e2e& 66.2 & 65.3 & 57.7 &\ 54.2 & 69.1 & 65.5 & 53.2 & 43.6 & 45.6 & 40.5 &56.7 & 43.9 \\
\hline
End-to-End & 62.7 &61.1 & 39.9 & 37.3 & 64.6 & 59.7 & 28.0 & 23.6 & 41.2 & 36.3 & 21.8 & 19.5\\
\hline
 \multicolumn{13}{c}{\textit{Multi-Domain Multi-Task Training}}  \\
\hline
ED-gold & -- & 95.0 & 65.2 & 62.5 & -- &  89.7 & 58.7 & 55.6 & -- & 86.3 & 78.1 & 71.0 \\

ED-pipeline & 68.1 & 66.9 & 58.8 & 55.3 & 70.7 & 69.0 & 51.5 &44.4 & \cellcolor{gray!40} 65.2 & \cellcolor{gray!40}64.6 & \cellcolor{gray!40}71.1 & \cellcolor{gray!40} 66.0 \\

ED-e2e & \cellcolor{gray!40}68.1 &\cellcolor{gray!40} 67.4 &\cellcolor{gray!40} 58.1 &\cellcolor{gray!40} 55.4 & \cellcolor{gray!40}72.2 &\cellcolor{gray!40} 69.3 & \cellcolor{gray!40}53.4 &\cellcolor{gray!40} 46.0 & 64.6 & 61.0 & 68.9 & 60.4\\
\hline
End-to-End & 63.9 &62.7 & 41.7 & 39.8 & 68.6 & 65.7 & 36.0 & 32.3 & 65.2 & 60.7 & 53.4 & 50.2\\
\hline

\multicolumn{13}{c}{\textit{Other Previous System (Single-Domain Training)}}  \\

\hline
OneIE  &   70.8   &    70.6 &     57.2     &  54.2         &  52.4             &       50.6  &           37.8   &        36.1     &      78.0        &      74.3    &           52.3   &              51.0  \\

DyGIE++   &  44.9  &  44.7  &     37.5  &      36.4     &      53.1  &           51.0       &        34.6   &   33.4   &              76.3            &       72.9  &                 62.7       &            60.5         \\

DEGREE(e2e)   &    60.9 & 60.7 & 36.0 & 27.0&  50.9  & 49.5   &   33.7   &   32.5   &   66.4        &   62.6  &      37.1   &     33.3      \\
DEGREE(pipe)   &      57.4 &   57.1 &   49.7   &  48.0       &   50.4  &  48.3  &    34.0   &   33.1       &    64.9      &               61.0  &         51.0     &          49.4           \\
TagPrime   &   69.5    & 69.3   & 63.3    &    61.0      &  71.7  & 71.1   &   60.9    &   51.7   &    75.7      &    73.0            &       61.8      &    60.8 \\          

\bottomrule

\end{tabular}
}
\caption{Event extraction performance (F1 \%) on CASIE, M2E2, and Genia2013 under single-domain and multi-domain multi-task training. The best scores are highlighted for each dataset. Multi-domain training consistently improves performance across datasets for the fully end-to-end mode. We also include comparisons with representative prior single-domain systems.}
\label{tab:f1_group1}
\end{table*}

\begin{table*}[h!]
\centering
\resizebox{\textwidth}{!}{
\begin{tabular}{l|cccc|cccc|cccc}
\toprule
 & \multicolumn{4}{c|}{RAMS} & \multicolumn{4}{c|}{Geneva} & \multicolumn{4}{c}{WikiEvents}  \\

Metrics & TI & TC & AI & AC & TI & TC & AI & AC & TI & TC & AI & AC  \\
\midrule
 \multicolumn{13}{c}{\textit{Single-Domain Multi-Task Training}} \\
\hline
ED-gold & -- & 37.0 & 53.5 & 46.9 & -- & 93.4 & 83.1 & 76.8 & -- & 72.3 & 64.1 & 58.0 \\
ED-pipeline & 79.6 & 30.7 & 46.8 & 34.9 &\cellcolor{gray!40} 84.6 &\cellcolor{gray!40} 81.4 & \cellcolor{gray!40}72.7 &\cellcolor{gray!40} 63.7 & 45.3 & 38.9 & 39.8 & 31.3 \\
ED-e2e& \cellcolor{gray!40}80.2 & \cellcolor{gray!40}30.6 &\cellcolor{gray!40} 46.7 &\cellcolor{gray!40} 34.8 & 83.4 & 81.2 & 72.1 & 62.8 &  &38.0 & 42.3 & 31.5 \\
\hline
End-to-End & 79.4 & 30.1 & 39.2 & 29.6 & 81.3 & 79.1 & 65.2 & 59.8 & 46.4 & 35.5 & 26.0 & 21.1 \\
\hline
 \multicolumn{13}{c}{\textit{Multi-Domain Multi-Task Training}} \\
\hline
ED-gold & -- & 38.9 & 53.0 & 45.8 & -- & 93.5 & 82.3 & 75.9 & -- & 79.6 & 64.9 & 59.9 \\

ED-pipeline & 79.1 & 28.9 & 45.7 & 34.4 & 84.6 & 81.4 & 71.8 & 61.6 & 46.6 & 41.4 & 39.8 & 32.4 \\

ED-e2e& 79.4 & 30.5 & 45.9 & 34.5 & 82.3 & 79.5 & 70.3 & 60.5 &\cellcolor{gray!40} 50.7 &\cellcolor{gray!40} 44.8 & \cellcolor{gray!40}42.4 &\cellcolor{gray!40} 33.4 \\
\hline
End-to-End & 78.9 & 32.1 & 39.4 & 30.6 & 80.7 & 78.7 & 64.8 & 59.5 & 49.3 & 43.3 & 28.9 & 24.2 \\
\hline

\multicolumn{13}{c}{\textit{Other Previous Systems (Single-Domain Training)}}  \\

\hline
OneIE(EAE)  &   --   &   --  &   48.0 & 40.7     &   --   &    --  &  38.9  &  37.1  &     --   &   --   & 17.5   &   15.0    \\

DyGIE++(EAE)   &  --    &   --   &   44.3   &  35.3   &           --   &     -- &     66.0   &   62.5  &    --   &    --  &     39.8  &    35.3      \\

DEGREE(EAE)   &   --   &    --  &   50.5 &   45.5      &    --  &   --   &   67.2 &   64.1          &    --  &    --  &     60.4     &  57.3      \\

TagPrime-C  &   --   & --   & 54.4   &    48.3      &  --  & --   &   83.0    &   79.2   &    --      &   --            &      70.4     &    65.7 \\

\bottomrule

\end{tabular}
}
\caption{F1 Scores for Event Extraction Tasks on RAMS, Geneva, WIKIEVENTS Datasets. Scores reported for previous systems  correspond only to argument extraction tasks using gold trigger words and types.}
\label{tab:f1_group2}
\end{table*}
\paragraph{Fine-grained Event Type Classification Challenges.} While coarse-grained, single-word types (e.g., \texttt{"Attack"}) offer simpler learning targets and higher data density, fine-grained, canonical types (e.g., \texttt{"Conflict\_Attack\_Detonate"}) introduce severe challenges of data sparsity. We plot the granularity distribution in Figure~\ref{fig:granularity} shown in Appendex~\ref{sec:granularity}. CASIE, Geneva, and Genia2013 operate primarily with coarse-grained (Levels 1-2) and moderately fine-grained (Level 3) event types and exhibit relatively stable performance. RAMS features numerous fine-grained event types (Levels 4-8) that suffer from data sparsity and require more nuanced semantic distinctions in event classification. On the RAMS dataset, the F1 scores for predicted event types matching at Level 1 (77.42\%) far exceed the event types matching at whole level (31.25\%) as presented in Table~\ref{tab:tc_offset_f1_levels}. Both WikiEvents and M2E2  include fine-grained event types, as shown in Figure~\ref{fig:granularity}; their annotations predominantly concentrate on the coarse-grained types. Analysis of granularity reveals that the complexity of multi-level event-type structures poses challenges that go beyond those caused by trigger word identification. The granularity of event types poses significant difficulties for proper annotation and model prediction.

\begin{table}[ht]
\centering
\begin{tabular}{lccc}
\hline
Dataset & Level 1 & Level 2 & Whole \\
\hline
M2E2 & 65.20 & -- & 63.74 \\
RAMS & 77.42 & 64.23 & 31.25 \\
WikiEvents  & 48.67 & 47.27 & 44.72 \\
\hline
\end{tabular}
\caption{Trigger Classification F1 (\%) decline by increasing granularity levels.}
\label{tab:tc_offset_f1_levels}
\end{table}

\paragraph{Misaligned in AI/AC Metrics.} \textbf{EAE}-related scores AI and AC of our systema and other previous systems presented in Table~\ref{tab:f1_group1} and Table~\ref{tab:f1_group2} demonstrate significant underperformance compared to \textbf{ED}-related scores. We observe systematic mismatches in evaluating the matching between the predicted argument text and gold annotation that reduce performance scores. For instance: the model correctly identifies the core argument element \texttt{Place=Boston}, but misses the possessive marker \texttt{"'s"} in the gold annotation \texttt{"Boston's"}. Similarly, the model correctly identifies the \texttt{victms=members}, but omits the determiner \texttt{'the'}, which is present in the gold span \texttt{"the members"}. This highlights a critical limitation of the metric: it prioritizes surface-form precision over semantic accuracy. Consequently, the reported scores for argument extraction, particularly for argument identification, are likely to significantly underestimate the model's true semantic understanding and extraction capability. Therefore, methods that improve the performance of the current system on EAE tasks, as well as a more appropriate evaluation metric in this field, require further exploration.

\paragraph{Summary.}
These results demonstrate that a unified multi-domain model achieves strong and competitive performance across datasets while significantly improving efficiency by removing the need for domain-specific models. Multi-domain training consistently benefits recall and generalization, particularly for resource-limited or schema-diverse datasets, highlighting effective cross-domain knowledge transfer. Moreover, fine-grained event type classification poses a significant challenge. Comparisons across ED-pipeline, ED-e2e, and fully end-to-end settings reveal that precise trigger identification remains critical. End-to-end modes offer more flexibility but tend to introduce additional spurious events and arguments. Notably, pipeline-based AI+AC benefits from explicit schema constraints, leading to more structurally consistent and controlled argument extraction compared to fully end-to-end approaches (see example predictions in Table~\ref{tab:full_geneva}). 
\section{Conclusion}
In this work, we demonstrate that a single T5-based Seq2Seq model based on unified multi-domain multi-task learning can effectively perform all event extraction subtasks across diverse domain datasets. Our results show that a unified model trained on consolidated multi-domain data achieves competitive performance compared to domain-specific models, while significantly improving efficiency by eliminating the need for multiple specialized systems. Notably, multi-domain training yields clear gains on resource-limited or schema-diverse datasets (e.g., Genia2013 and WIKIEVENTS), highlighting the benefits of cross-domain knowledge transfer for improved generalization.

However, fully end-to-end inference lacks explicit control over structural and role constraints, and is further limited by the relatively scarce and less diverse supervision signals available for such formulations. Additionally, the fine-grained granularity of event types poses challenges for both annotation consistency and model prediction. While the proposed unified framework provides a practical and scalable solution for multi-domain information extraction, maintaining sufficient domain similarity remains important to ensure stable performance. Overall, our unified framework offers a strong balance between performance and efficiency, providing a streamlined and adaptable framework for real-world deployment across heterogeneous domains. Future work will focus on improving robustness to fine-grained event structures, as well as exploring more flexible learning strategies to better accommodate domain-specific linguistic variations.

\section*{Limitations}
Our work has several limitations that present opportunities for future research. Firstly, the scope of our cross-domain evaluation was limited by the availability of event extraction datasets from various domains, which may impact the broader generalizability of our findings. Secondly, we deliberately restricted our focus to event extraction tasks. Although the unified model architecture is extensible in principle to broader information extraction tasks, such as entity and relation extraction, a systematic investigation of this extension remains a focus for future work. Thirdly, the emergence of powerful large language models (LLMs) introduces a compelling alternative paradigm for generative information extraction, particularly in multi-domain settings where unified training can leverage broad contextual knowledge. A systematic comparison between unified multi-domain models and LLM-based approaches, especially in terms of efficiency and scalability, remains an important direction for future work.
\section*{Acknowledgment}
We greatly appreciate the foundational work of \textbf{TEXTEE} \cite{Huang23textee}, which undertook the significant effort to benchmark existing event extraction systems and standardize the preprocessing of diverse datasets. 

\section*{Declaration on  the Use of AI Assistants}
During the preparation of this work, we acknowledge the use of GPT-4o-mini for only spell checking, paraphrasing, and latex formatting purposes; Copilot for reviewing coding errors. After the use of the AI tools, We systematically reviewed and edited the all content as needed and take full responsibility for the publication’s content.



\bibliography{custom}

\appendix

\section{Comparison of capabilities of different existing EE systems}
\label{sec:appendix_a}
presents a comparative overview of representative event extraction (EE) systems across key sub-tasks, as well as their support for cross-domain generalization. The systems span two main paradigms: classification-based approaches (e.g., DyGIE++, OneIE, TagPrime) and generation-based methods (e.g., DEGREE, BartGEN). Our approach extends the generation paradigm by adopting a unified sequence-to-sequence (Seq2Seq) framework that jointly models all EE components while explicitly supporting cross-domain transfer. This enables consistent handling of diverse event schemas and improves scalability compared to prior systems.

\section{Comparisions of Different Model Architectures.}
\label{sec:preliminary}
\begin{table}[h]
\centering
\resizebox{\columnwidth}{!}{
\begin{tabular}{lcccc}
\hline
\textbf{Model} & \textbf{TI} & \textbf{TC} & \textbf{AI} & \textbf{AC} \\
\hline
T5-base    & 84.6 & 81.4 & 72.7 & 63.7 \\
GPT-2      & 58.1 & 48.2 & 44.8 & 27.4 \\
BART       & 82.5 & 70.3 & 55.5 & 55.4 \\
T5-large   & 85.1 & 81.9 & 70.7 & 59.3 \\
\hline
\end{tabular}
}
\caption{F1 scores (\%) on the Geneva dataset for trigger identification (TI), trigger classification (TC), argument identification (AI), and argument classification (AC).}
\label{tab:geneva_results}
\end{table}
We compare the generation behaviors of T5-base, BART, and GPT-2 on event extraction in preliminary experiments. The results in Table~\ref{tab:geneva_results} reveal clear performance gaps across model architectures, particularly between encoder–decoder models (T5, BART) and the decoder-only GPT-2. Trigger identification indicates that trigger detection is relatively robust for encoder-decoder models. However, substantial differences emerge in argument extraction quality. Given the example of successfully identifying the event trigger (“pressing”) shown in Table~\ref{tab:model_generations}, GPT-2, as a decoder-only model, exhibits severe over-generation, including repeated structures, inconsistent argument-role assignments, and formatting errors, reflecting its weaker control over structured outputs. BART, despite its encoder–decoder architecture, shows improved coherence but still introduces minor inconsistencies in argument alignment and formatting, likely due to differences in pretraining objectives and less effective conditioning on structured prompts compared to T5. In contrast, T5-base produces fully correct and well-structured predictions that closely match the gold reference. Its text-to-text formulation and strong prompt adherence enable precise argument identification and role assignment. Overall, while trigger prediction is consistent across models, argument extraction highlights clear architectural differences, with T5 demonstrating superior reliability and structural fidelity.

\section{Zero-shot prompting for ED-e2e.}
\label{sec:gpt4o}
We evaluate GPT-4o-mini under a zero-shot ED-e2e setting using a constrained prompting template. Given an input sentence and a set of candidate event types, the model is instructed to directly generate trigger-type pairs in a structured format:
\textit{trigger > $\langle$trigger\_text$\rangle$ $=>$ $\langle$EventType$\rangle$}. The prompt explicitly enumerates all valid event labels and enforces label selection from this predefined set. Despite this constraint, the results in Table~\ref{tab:gpt4o-mini} show that GPT-4o-mini struggles with accurate trigger localization and classification. This suggests that while label-constrained prompting can partially guide type prediction, it is insufficient to ensure precise span detection or consistent alignment with gold triggers. Consequently, the low TI/TC performance indicates that directly extending this setup to argument extraction would be unreliable. 
\begin{table}[ht]
\centering
\small
\setlength{\tabcolsep}{3pt}
\resizebox{\columnwidth}{!}{
\begin{tabular}{l|cccccc}
\toprule
\textbf{F1} & CASIE &M2E2 & Genia2013 & RAMS & Geneva & WikiEvents \\
\midrule
\textbf{TI} & 5.96  & 40.79 & 11.57 & 6.28 & 5.64 & 17.58 \\
\textbf{TC} & 5.63  & 34.21 & 8.76  & 1.96 & 1.98 & 12.68 \\
\bottomrule
\end{tabular}
}
\caption{Trigger identification (TI) and trigger classification (TC) F1 (\%) scores for GPT-4o-mini under zero-shot ED-e2e setting across six datasets.}
\label{tab:gpt4o-mini}
\end{table}

\section{Benefit of Multi-domain Learning.}
\label{app:multi-domain learning}
Joint multi-domain training exhibits a consistent divergence between precision and recall dynamics across datasets. Precision tends to stabilize early and remains largely unchanged throughout training, particularly in high-resource or simpler domains such as Geneva. Recall improves progressively with training, with the most substantial gains observed in challenging and heterogeneous domains, indicating effective cross-domain knowledge transfer.

\section{Granularity Distribution of different Domains.}
\label{sec:granularity}
\begin{figure*}[t]
\centering
\resizebox{\textwidth}{!}{
    \includegraphics[width=\textwidth]{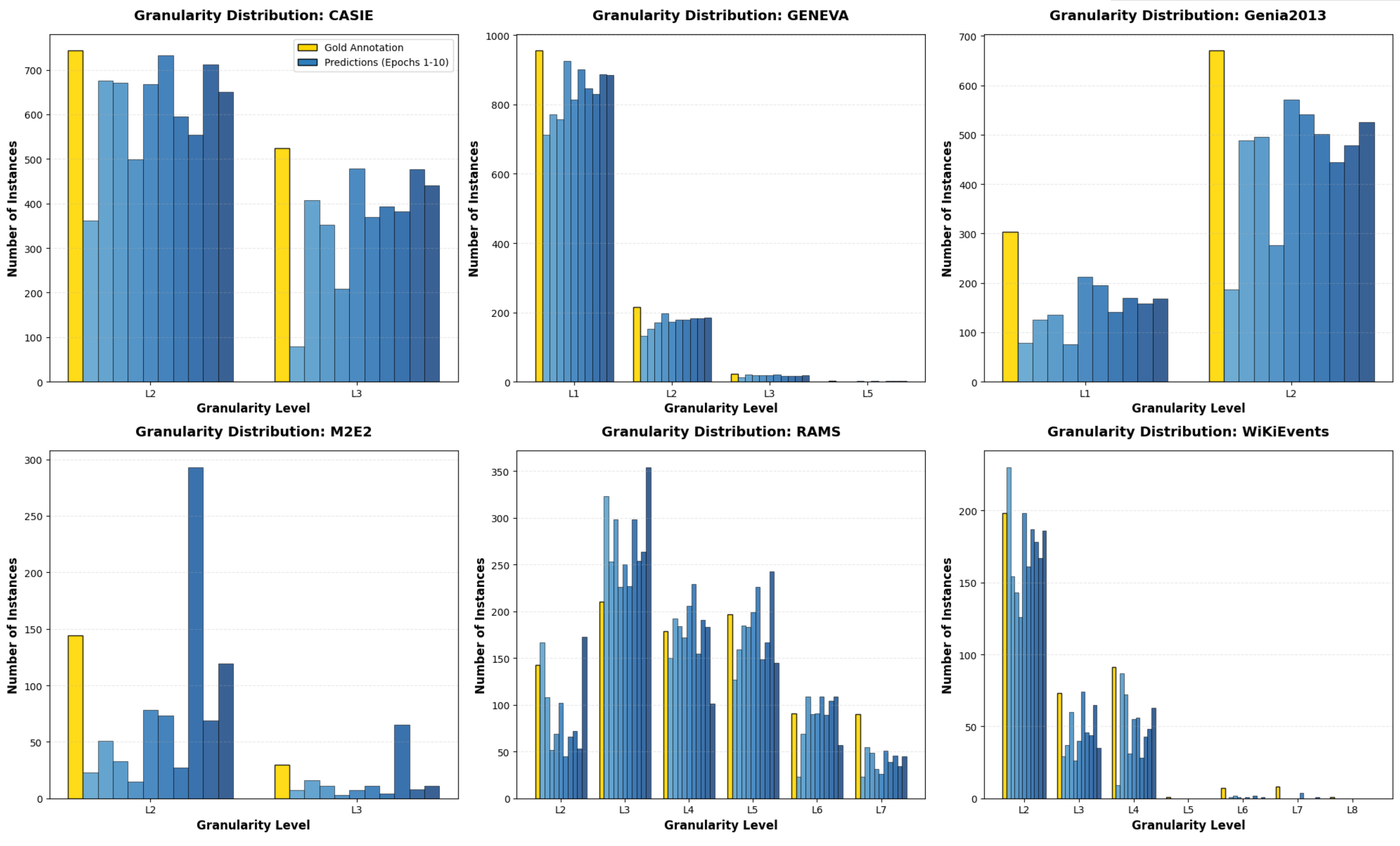}
    }
    \caption{Comparison of event type granularity distributions of different datasets between annotation and predictions across training Epochs.}
    \label{fig:granularity}
    
\end{figure*}

Figure \ref{fig:granularity} presents the granularity distribution of each dataset. The analysis of granularity distribution reveals a fundamental divergence that directly impacts model performance in trigger classification (TC). Because granularity introduces significant challenges, including data sparsity for rare event types and more complex semantic decision boundaries, which consequently lead to higher error rates and explain the performance gap observed across the diverse datasets. In particular, datasets with deeper and more uneven granularity hierarchies (such as RAMS and WikiEvents), where fine-grained event types account for a disproportionately small portion of the data, are more difficult to learn effectively. By contrast, datasets with shallower or more balanced granularity levels (e.g., CASIE and Genia2013) allow models to achieve more stable predictions, highlighting the importance of granularity-aware modeling strategies for robust cross-domain generalization.



\begin{table*}[ht]
\centering
\resizebox{\textwidth}{!}{
\begin{tabular}{lcccc|c}
\toprule
\textbf{Model} & \textbf{TI} & \textbf{TC} & \textbf{EAE} & \textbf{Cross-Domain}  & \textbf{Paradigm} \\
\midrule

DEGREE \cite{Hsu22degree} & \checkmark & \checkmark & \checkmark & -- &  Generation \\

BartGEN\cite{Li21bartgen} & -- & -- & \checkmark & -- & Generation \\

DyGIE++ \cite{Wadden19dygiepp} & \checkmark & \checkmark & \checkmark & -- & Classification \\
OneIE\cite{Lin20oneie} & \checkmark & \checkmark & \checkmark & -- & Classification \\

TagPrime\cite{Hsu23tagprime} & -- & \checkmark & \checkmark & -- & Classification (Seq Tagging) \\
Ours & \checkmark & \checkmark & \checkmark  & \checkmark & Generation (Seq2Seq) \\
\bottomrule
\end{tabular}
}
\caption{Coverage of event extraction tasks and paradigm for each model. }
\label{tab:ee-tasks-paradigm}
\end{table*}

\begin{table*}[h]
\centering

\begin{tabular}{|l|p{\dimexpr\textwidth-4cm\relax}|}
\hline
\textbf{Model} & \textbf{Prediction} \\
\hline
\textbf{Gold Reference} & 
trigger> pressing>: argument> Iran>=Addressee | argument> to give up its nuclear program altogether>=Content | argument> the US leader>=Speaker | None=Topic \\
\hline
\textbf{T5 Base} & 
trigger> pressing>: argument> Iran>=Addressee | argument> to give up its nuclear program altogether>=Content | argument> the US leader>=Speaker | None=Topic \\
\hline
\textbf{BART Base} & 
trigger> pressing>: argument> Iran>=Addressee | argument> to give up its nuclear program altogether>=Content | \textcolor{red}{the US leader>=Speaker} | None=Topic \\
\hline
\textbf{GPT-2} & 
trigger> pressing>: argument> Iran>=Addressee | argument> to give up its nuclear program altogether>=Content | argument> Sharon>=Speaker | None=Topic>=Topic \textcolor{red}{for event Convincing of trigger> pressing>: None=Addressee | argument> the US leader>=Content | argument> to keep>=Speaker | None=Topic>=Topic for event Convincing of trigger> pressing>: None=Addressee | argument> Iran>=Content | argument> to give up its nuclear program altogether>=Addressee | argument} \\
\hline
\end{tabular}
\caption{Event Extraction Predictions: Event with Trigger \textit{pressing}. GPT-2 struggles due to being decoder-only, producing over-generation, repetition, and label errors. BART is better but still less precise than T5, especially in formatting and alignment with prompts. T5’s encoder–decoder design and text-to-text pretraining allow accurate, structured, prompt-compliant generation, making it the most suitable base model for unified event extraction.}
\label{tab:model_generations}
\end{table*}

\begin{figure*}[h]
\centering
\begin{subfigure}{\textwidth}
    \centering
    \includegraphics[width=\textwidth]{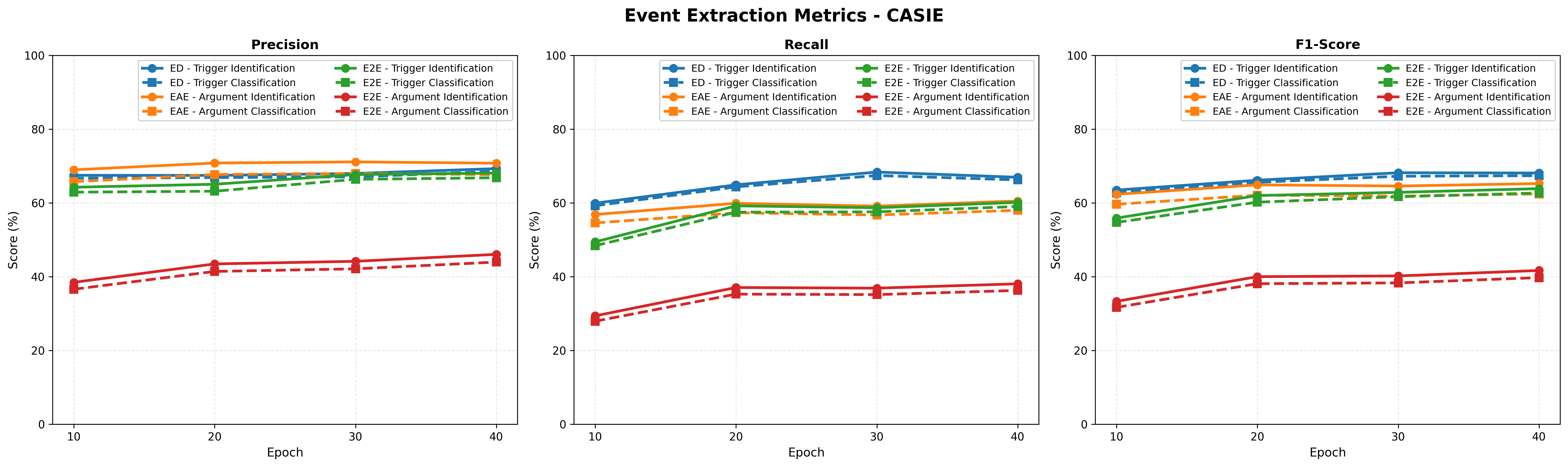}
    \caption{Training dynamics across epochs. Each row shows precision, recall, and F1-score curves over epochs for ED, EAE, and E2E paradigms on the CASIE dataset. }
    \label{fig:epochs_casie}
\end{subfigure}

\vspace{2mm}

\begin{subfigure}{\textwidth}
    \centering
    \includegraphics[width=\textwidth]{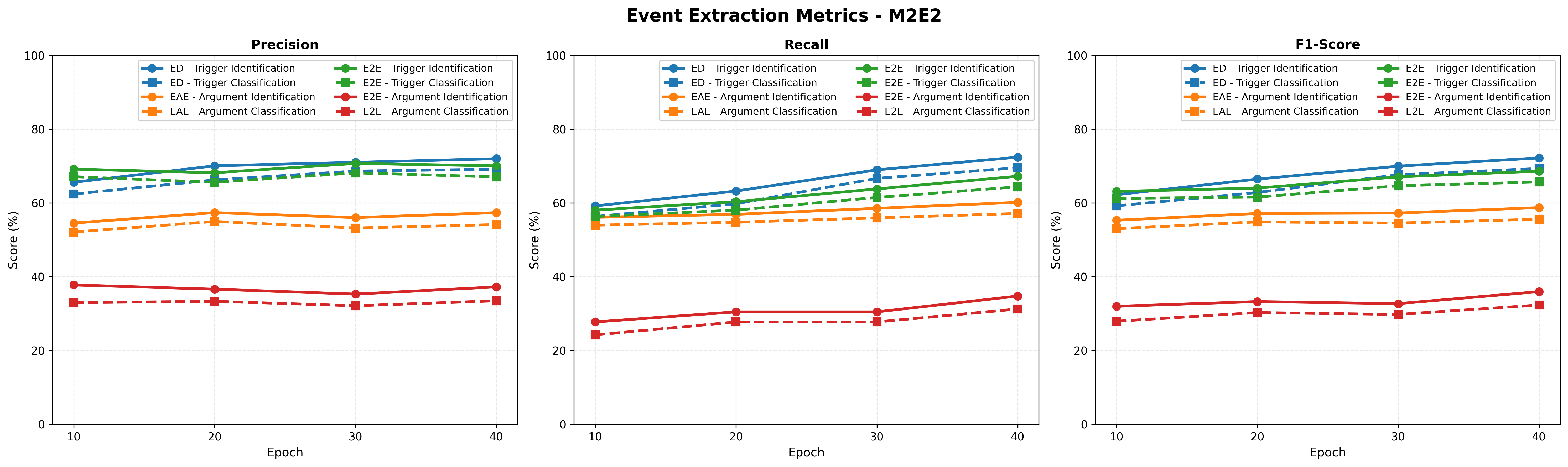}
    \caption{Training dynamics across epochs. Each row shows precision, recall, and F1-score curves over epochs for ED, EAE, and E2E paradigms on the M2E2 dataset.}
    \label{fig:epochs_m2e2}
\end{subfigure}

\vspace{2mm}

\begin{subfigure}{\textwidth}
    \centering
    \includegraphics[width=\textwidth]{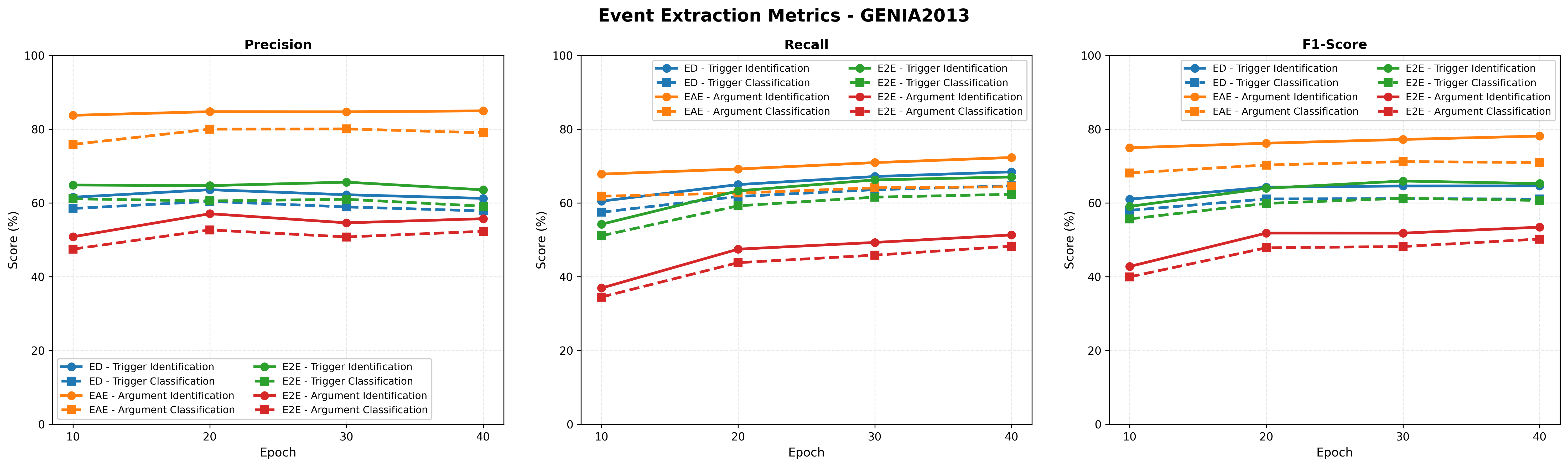}
    \caption{Training dynamics across epochs. Each row shows precision, recall, and F1-score curves over epochs for ED, EAE, and E2E paradigms on the Genia2013 dataset.}
    \label{fig:epochs_genia2013}
\end{subfigure}
\end{figure*}

\begin{figure*}[h]
\begin{subfigure}{\textwidth}
    \centering
    \includegraphics[width=\textwidth]{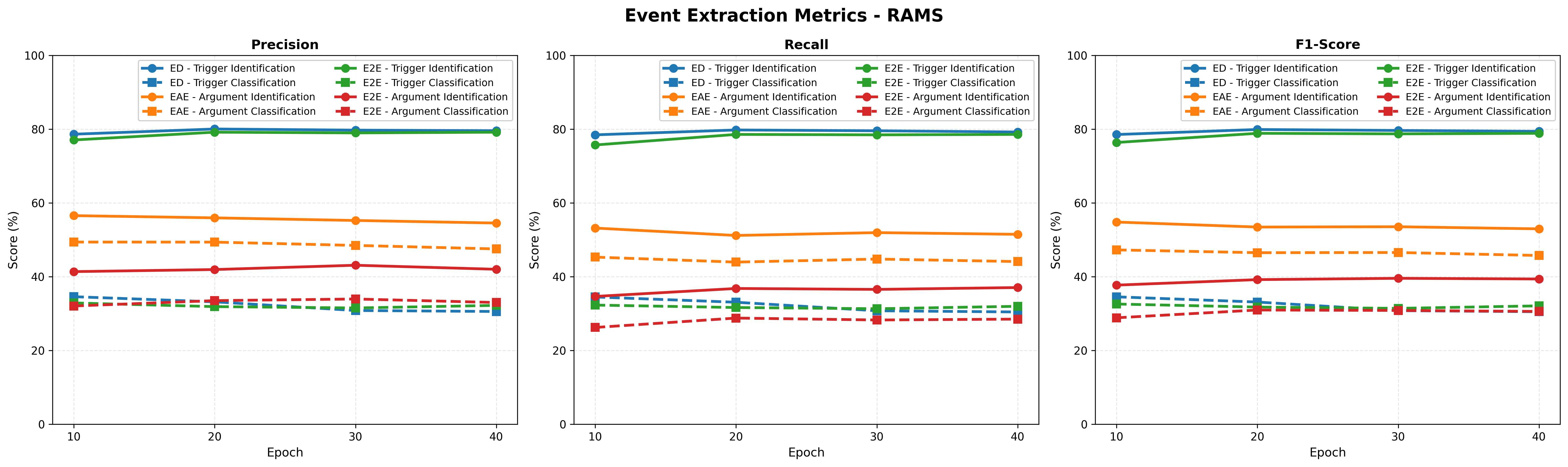}
     \caption{Training dynamics across epochs. Each row shows precision, recall, and F1-score curves over epochs for ED, EAE, and E2E paradigms on the RAMS dataset.}
    \label{fig:epochs_rams}
\end{subfigure}

\vspace{2mm}

\begin{subfigure}{\textwidth}
    \centering
    \includegraphics[width=\textwidth]{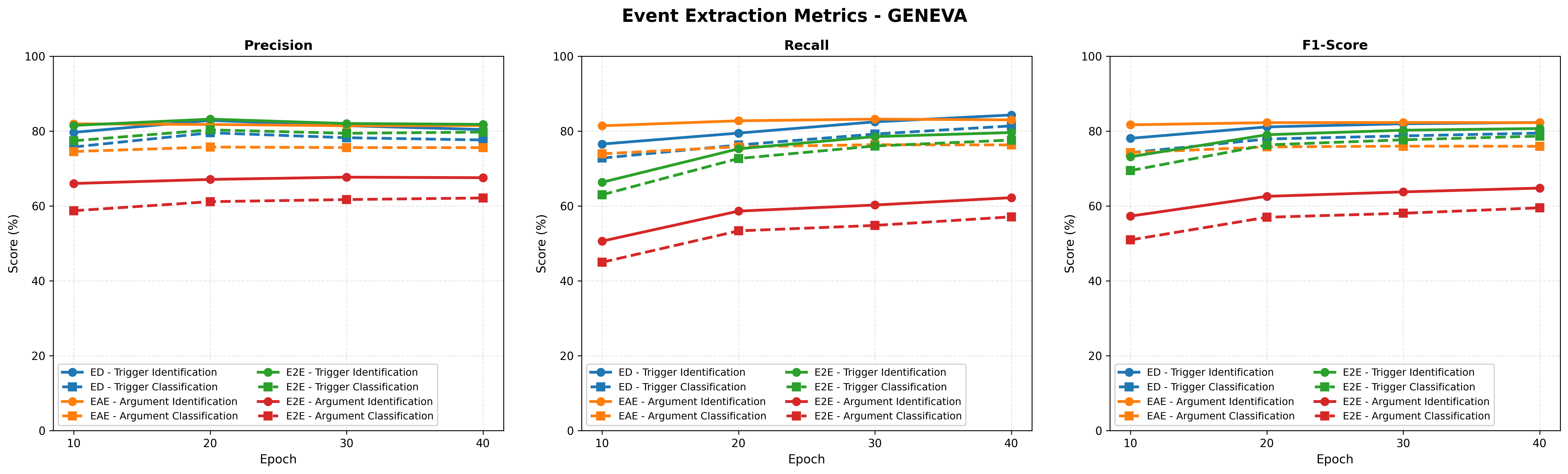}
     \caption{Training dynamics across epochs. Each row shows precision, recall, and F1-score curves over epochs for ED, EAE, and E2E paradigms on the Geneva dataset.}
    \label{fig:epochs_geneva}
\end{subfigure}

\vspace{2mm}

\begin{subfigure}{\textwidth}
    \centering
    \includegraphics[width=\textwidth]{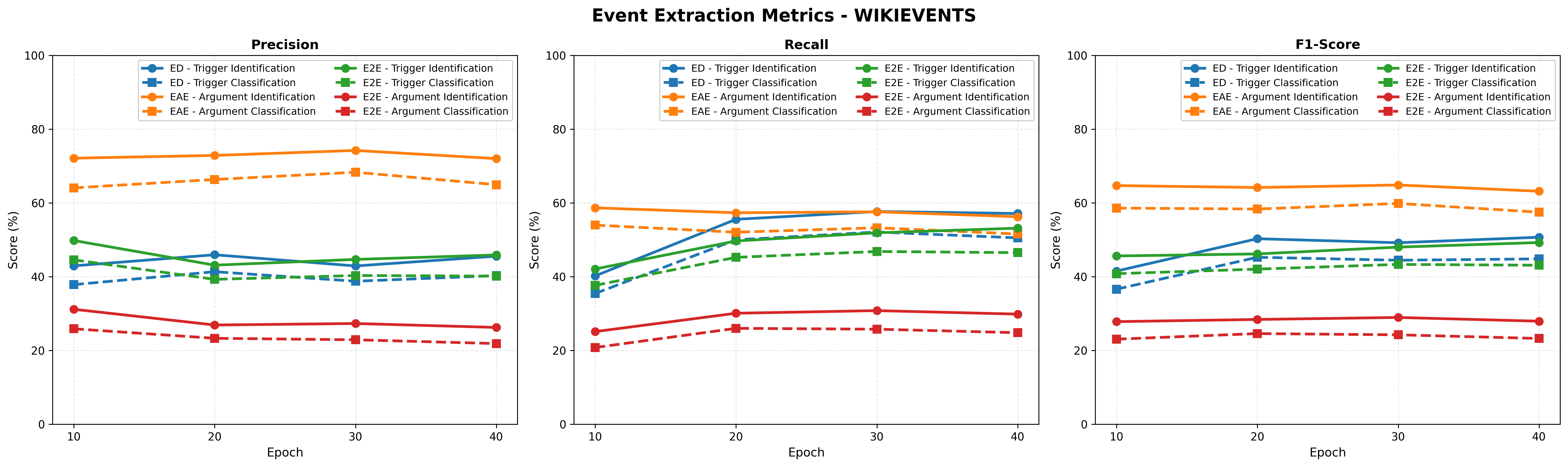}
     \caption{Training dynamics across epochs. Each row shows precision, recall, and F1-score curves over epochs for ED, EAE, and E2E paradigms on the WikiEvents dataset.}
    \label{fig:epochs_wikievents}
\end{subfigure}
\end{figure*}

\begin{table*}[t]
\centering
\scriptsize
\setlength{\tabcolsep}{2pt}
\renewcommand{\arraystretch}{1.1}

\begin{tabularx}{\textwidth}{
l
>{\raggedright\arraybackslash}X 
>{\raggedright\arraybackslash}X 
>{\raggedright\arraybackslash}X 
>{\raggedright\arraybackslash}X 
>{\raggedright\arraybackslash}X}
\toprule
& & \multicolumn{2}{c}{\textbf{Single-Domain}} &\multicolumn{2}{c}{\textbf{Multi-Domain}}\\
\midrule
\textbf{Sub-Task} & \textbf{Reference} 
& \textbf{Input (Single)} & \textbf{Prediction (Single)} 
& \textbf{Input (Multi)} & \textbf{Prediction (Multi)} \\
\midrule

TI
& When farmers became capable of producing food beyond the needs of their own families, others in their society were freed to devote themselves to projects other than food trigger>acquisition>.
& Domain: geneva. Identify trigger words. When farmers became capable of producing food beyond the needs of their own families, others in their society were freed to devote themselves to projects other than food acquisition.
& When farmers trigger>became> capable of trigger>producing> food beyond the needs of their own families, others in their society were freed to devote themselves to projects other than food trigger>acquisition>.
& Domain: geneva. Identify trigger words. When farmers became capable of producing food beyond the needs of their own families, others in their society were freed to devote themselves to projects other than food acquisition.
& When farmers trigger>became> capable of trigger>producing> food beyond the needs of their own families, others in their society were freed to devote themselves to projects other than food trigger>acquisition>. \\

\midrule

ED-pipeline
& acquisition=Getting
& Domain: geneva. Classify trigger types. When farmers trigger>became> capable of trigger>producing> food beyond the needs of their own families, others in their society were freed to devote themselves to projects other than food trigger>acquisition>.
& became$=$Becoming | \newline producing$=$Manufacturing | \newline acquisition$=$Getting
& Domain: geneva. Classify trigger types. When farmers trigger>became> capable of trigger>producing> food beyond the needs of their own families, others in their society were freed to devote themselves to projects other than food trigger>acquisition>.
& became$=$Becoming | \newline producing$=$Manufacturing | \newline acquisition$=$Getting \\

\midrule

ED-e2e
& trigger>acquisition>\newline=Getting
& Domain: geneva. Classify trigger types end-to-end. When farmers became capable of producing food beyond the needs of their own families, others in their society were freed to devote themselves to projects other than food acquisition.
& trigger>became>\newline=Becoming | trigger>producing>\newline=Manufacturing | trigger>freed>\newline=Causation | trigger>acquisition>\newline=Getting
& Domain: geneva. Classify trigger types end-to-end. When farmers became capable of producing food beyond the needs of their own families, others in their society were freed to devote themselves to projects other than food acquisition.
& trigger>became>\newline=Becoming | trigger>producing>\newline=Manufacturing | trigger>acquisition>\newline=Getting \\

\midrule

EAE(ED-pipeline)
& trigger>acquisition>\newline=Getting: None=Means | None=Recipient | None=Source | \newline argument>food>=Theme
 & Domain: geneva. Extract arguments for a specific event. argument> Entity | Final\_category | Final\_quality for trigger> became>=Becoming; argument> Factory | Instrument | Producer | Product | Resource for trigger> producing>=Manufacturing; argument> Means | Recipient | Source | Theme for trigger>acquisition>\newline=Getting: When farmers became capable of producing food beyond the needs of their own families...
&  trigger> became>=Becoming: argument> farmers>=Entity | None=Final\_category | argument> capable of producing food beyond the needs of their own families>=Final\_quality; trigger> producing>\newline=Manufacturing: None=Factory | None=Instrument | argument> farmers>=Producer | argument> food beyond the needs of their own families>=Product | None=Resource; trigger>acquisition>\newline=Getting: None=Means | None=Recipient | None=Source |\newline argument>food>=Theme
& Domain: geneva. Extract arguments for a specific event. argument> Entity | Final\_category | Final\_quality for trigger> became>=Becoming; argument> Factory | Instrument | Producer | Product | Resource for trigger> producing>=\newline Manufacturing; argument> Means | Recipient | Source | Theme for trigger>acquisition>\newline=Getting: When farmers became capable of producing food beyond the needs of their own families...
& trigger> became>=Becoming: argument> farmers>=Entity | None=Final\_category | argument> capable of producing food beyond the needs of their own families>=Final\_quality; trigger> producing>=Manufacturing: None=Factory | None=Instrument | argument> farmers>=Producer | argument> food>=Product | None=Resource; trigger> acquisition>=Getting: None=Means | None=Recipient | None=Source | argument> food>=Theme \\

\midrule
End-to-End
& trigger>acquisition>\newline=Getting | argument>food>=Theme
& Domain: geneva. Event trigger and argument extraction. When farmers became capable of producing food beyond the needs of their own families, others in their society were freed to devote themselves to projects other than food acquisition.
& trigger>became>\newline=Becoming | argument>farmers>\newline=Entity | argument>capable of producing food beyond the needs of their own families>\newline=Final\_category $||$ trigger>producing>\newline=Manufacturing | argument>farmers>\newline=Producer | argument>food>\newline=Product $||$ trigger>freed>=Causation | argument>others in their society>\newline=Affected | argument>to devote themselves to projects other than food acquisition>\newline=Effect $||$ trigger>acquisition>\newline=Getting | argument>food>=Theme
& Domain: geneva. Event trigger and argument extraction. When farmers became capable of producing food beyond the needs of their own families, others in their society were freed to devote themselves to projects other than food acquisition.
& trigger>became>\newline =Becoming| argument>farmers>\newline=Entity | \newline argument>capable of producing food beyond the needs of their own families>\newline=Final\_quality$||$\newline trigger>producing>\newline=Manufacturing | argument>farmers>\newline=Producer| argument>food>\newline=Product $||$\newline trigger>acquisition>\newline=Getting|argument>food>\newline=Theme \\

\bottomrule
\end{tabularx}

\caption{Comparison of prediction per sub-task between single-domain and multi-domain models on Geneva.}
\label{tab:full_geneva}
\end{table*}

\end{document}